# Improving LLM Abilities in Idiomatic Translation

**Sundesh Donthi[†], Maximilian Spencer[†*], Om Patel[†]**
Joon Young Doh[†], Eid Rodan[†], Kevin Zhu[†], Sean O'Brien[†]
[†]Algoverse AI Research
{kevin, sean}@algoverse.us

## Abstract

Translating idiomatic expressions remains a challenge for large language models (LLMs), as they often produce literal, semantically incorrect translations—for instance, directly converting "break a leg" into a nonsensical phrase in the target language. While external resources like IdiomKB can supply the figurative meaning and thus yield semantically accurate translations, this approach does not preserve the cultural and stylistic nuances that make idioms so distinctive. Our study focuses on idiomatic translations across multiple languages, including Chinese (ZH), Urdu (UR), and Hindi (HI), with clearly defined abbreviations for each. We propose two methods for improving idiomatic translation fidelity: a Semantic Idiom Alignment (SIA) approach that uses pre-trained sentence embeddings to identify target-language idioms, and a Language-Model-based Idiom Alignment (LIA) approach that prompts an LLM to suggest appropriate idiom counterparts. Human evaluations across multiple language pairs show that SIA better preserves idiomatic style. To support this work, we introduce idiom datasets in low-resource languages (Urdu and Hindi). Our results indicate that aligning idioms at the semantic level can improve cross-lingual style preservation and cultural authenticity. All resources created can be found at: github.com/mzx05/IdiomaticTranslationResearch

## 1 Introduction

Global communication increasingly relies on machine translation, yet current large language models (LLMs) often fail to preserve the cultural and emotional nuances inherent in idiomatic expressions. Idioms are linguistically and culturally bound, and translating them accurately is crucial for maintaining the original text's authenticity and resonance. While recent work has augmented LLMs like NLLB and GPT with external knowledge bases (e.g., IdiomKB) to improve semantic correctness, these approaches do not retain idiomatic style and contextual richness.

[*]Corresponding Author: mspencer2@binghamton.edu

Figure 1: This is an example of idiomatic translation from Chinese to English. Since this idiom's literal translation is similar to the figurative meaning, the direct translation still conveys the proper meaning. However, both the SIA and LIA methods are favorable as they maintain idiomatic style in the translation.

This gap motivates our central research question: How can we enhance LLM-based translation to capture both the semantic content and the culturally rooted idiomatic flair of the source language? We address this by developing two complementary strategies. First, we propose a Semantic Idiom Alignment (SIA) method that leverages pre-trained sentence embeddings (e.g., Sentence Transformers) and cosine similarity measures to identify target-language idioms closely aligned with the source-language meaning. Specifically, we embed the English meaning of each idiom across multiple languages, retrieve top-K candidates via cosine similarity, and then prompt GPT4o to select the most culturally and contextually appropriate match. Second, we introduce a LLM-based Idiom Alignment (LIA) approach, prompting the LLM itself to propose suitable idiomatic counterparts directly.

To support these methods, we curate expanded idiom datasets, including low-resource Urdu and Hindi idioms indexed by their English meanings. We evaluate our approaches through both human judgments and model-based assessments, finding that SIA, in particular, improves idiomatic fidelity and cultural nuance across multiple language pairs.

Our contributions are summarized as follows: (1)

introducing methods designed specifically for cross-lingual idiom preservation, (2) constructing benchmark datasets for Urdu and Hindi idioms, and (3) demonstrating that enhancing LLM-based translations with idiom alignment can significantly improve stylistic and cultural authenticity. This work lays a foundation for more nuanced and culturally aware machine translation models, ultimately enabling richer, more faithful global communication.

## 2 Background

### 2.1 Limitations in Machine Translation of Idioms

From a literary standpoint, idioms are figurative, institutionalized expressions that enrich speech and writing, demonstrating mastery of a language. Language models must understand and interpret idioms, especially when translating from one language to another. Recent work has used IdiomKB as a knowledge base for translating idioms, achieving some success with language models (Li et al.,2023). This knowledge base pairs idioms in a language with their meanings in English, Chinese, and Japanese. In their method, they use this to provide the translation model with the figurative meaning of the idiom in the sentence, resulting in more semantically correct translations. However, the knowledge base is limited to only three languages and it does not include any low-resource languages, and the translations do not maintain an idiomatic expression.

Building on these techniques for idiomatic translation is the use of retrieval-augmented models (KNN-MT) and the upweighting of training loss on potentially idiomatic sentences (Liu and Neubig, 2023). This showed improvements in translations for idiomatic sentences along with slight improvements in non-idiomatic sentences as well. However, limitations include the use of synthetic data, limited languages, and the heavy reliance on high-quality training data. Past research has focused on translating an idiom in the original language to the figurative meaning in the target language. Although this may convey the message, it fails to be a true translation because the idiomatic sentence style is lost.

### 2.2 Next Steps to Build On IdiomKB

As evidenced by Li and Chen, the use of specialized knowledge bases such as IdiomKB has proven beneficial. However, the limited scope of these resources, covering only a few languages, constrains their utility in broader linguistic contexts (Li et al.,2023). This highlights the need to expand these databases to encompass a wider array of languages and idiomatic expressions. We also hope to build on the use of a knowledge base in idiomatic translation by using it to translate an idiom in one language to an idiom in another language. This would better capture the meaning of the sentence and help maintain the style of the idiomatic sentence across languages. The inherent complexity of idioms is underscored by research from Dankers and Lucas, who analyze the compositional challenges faced by Transformer models in handling idiomatic expressions. Their findings reveal that while these models adeptly process standard grammatical constructions, they frequently misinterpret the non-compositional nature of idioms, leading to incomplete or incorrect translations (Dankers et al.,2022). This suggests that current models need enhancements in semantic flexibility to better accommodate the abnormalities of idiomatic language. Further highlighting the translation challenges, Shao and Sennrich's evaluation of machine translation performance on idiomatic texts points out that even advanced models struggle to maintain the expressive depth and cultural nuances of idioms, often resulting in translations that are either too literal or misleading (Shao et al., 2017). The necessity for more refined training datasets specifically tailored to improve the handling of idiomatic expressions within translation systems becomes an emphasized need after understanding the limitations of such technology.

### 2.3 Newer Idiom Knowledge Resources

In response to these challenges, new resources such as the EPIE dataset introduced by Saxena and Paul are emerging. This dataset aims to enhance the identification and translation of idiomatic expressions by providing context-rich examples of their usage across various languages (Saxena and Paul, 2020). Such resources are invaluable for developing more sophisticated models capable of recognizing and translating idioms accurately. The work of Liu et al. offers a promising direction through the application of retrieval-augmented models and idiomatic sentence-focused training techniques. Their approach shows improvements in translating idiomatic sentences and enhances the overall fluency of translated texts, suggesting a viable pathway to overcome some inherent limitations of current translation models (Liu and Neubig, 2023).

Table 1: Limitations of Previous Research in Idiomatic Translation

| Study | Key Limitations |
|---|---|
| Li et al. (2023) | Limited language support (only three languages); does not cover low-resource languages; translation does not maintain an idiomatic expression |
| Liu et al. (2023) | Heavy reliance on synthetic data; models show only slight improvements in non-idiomatic sentences; limited coverage of idiomatic expressions |
| Dankers and Lucas (2022) | Transformer models fail to process non-compositional idiomatic expressions accurately, leading to incomplete or incorrect translations |
| Shao and Sennrich (2017) | Advanced models struggle to maintain cultural nuances and expressiveness of idioms; translations are often too literal or misleading |
| Saxena and Paul (2020) | Emerging dataset but lacks comprehensive coverage across languages and idiomatic variations |

### 2.4 Addressing the Limitations of Previous Research

The limitations outlined in Table 1 reveal several gaps in the current approaches to idiomatic translation. Li et al. introduced a knowledge base, IdiomKB, which provides figurative meanings for idioms, but its limited language support and exclusion of low-resource languages restrict its applicability in broader contexts. Similarly, Liu et al. (Liu and Neubig, 2023) made improvements in translating idiomatic sentences through retrieval-augmented models, but the heavy reliance on synthetic data and the minimal improvements in non-idiomatic sentences highlight the need for more comprehensive and natural training datasets.

Furthermore, Dankers and Lucas (Dankers et al.,2022) pointed out the challenge that Transformer models face in processing non-compositional idioms, often resulting in incomplete or incorrect translations. Shao and Sennrich (Shao et al., 2017) also noted that even advanced models struggle to maintain cultural nuances and expressiveness, often producing overly literal translations. While Saxena and Paul (Saxena and Paul, 2020) introduced the EPIE dataset to enhance idiomatic expression translation, its limited coverage of languages and idiomatic variations underscores the need for more expansive resources.

Our proposed research aims to address these limitations by expanding the range of supported languages, particularly focusing on low-resource languages. Additionally, we introduce a novel approach that not only translates idioms but also preserves the idiomatic style and cultural nuances across languages. By building on existing models and incorporating a refined, context-rich dataset, our approach seeks to improve both the accuracy and cultural fidelity of idiomatic translations across diverse linguistic contexts.

## 3 Method

### 3.1 Dataset construction

For the English-to-Chinese translation, we used the "MWE-PIE" (Zhou and Bhat, 2021) dataset that had 1,197 English idioms with around 5 sentences per idiom for a total of 5,170 sentences. For the Chinese-to-English translation, we used the CCT "cheng yu" dataset (Tan, 2021) which had 108,987 Chinese sentences that contained 7,397 unique Chinese idioms. We utilized two files shared by the IdiomKB team. The datasets had the following attributes: an id for indexing, an idiom, English meaning, and the Chinese meaning(in that order). For the Urdu dataset construction, we found a dataset with 2,111 Urdu idioms (with repeats) (Hussain et al., 2021) and their English meanings/idioms. We then found matching English idioms when they existed from our English idiom dataset and, using GPT4o, generated English sentences for those that we did not already have sentences that we flagged. For the Hindi dataset construction, we manually compiled 990 Hindi idioms, Hindi meanings, and Hindi sentences from reputable websites, ensuring there are no duplicates. We generated the English meanings for these idioms from the Hindi meanings using GPT-4o.

To facilitate future use with the SIA method, we restructured the datasets so the English meaning serves as the key, with the meanings and idioms from other languages as the values. We are indexing on the English meanings so that semantically comparing the English meanings of idioms is made easier (Li et al.,2023). For the purpose of our research, the idiomatic knowledge bases are exhaustive enough because across languages there were an adequate number of idioms that found a match in the target language.

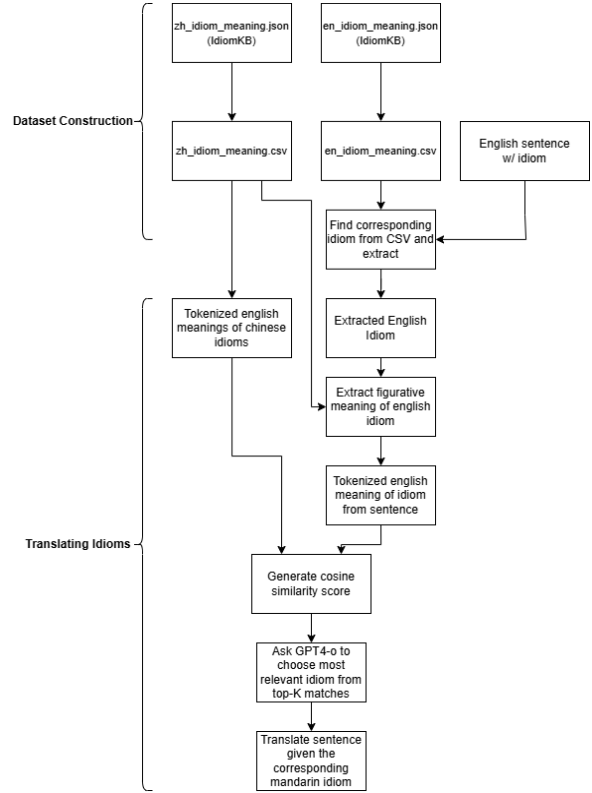

Figure 2: Dataset Construction and Translation Idioms (SIA) Flow Chart

### 3.2 Translating Idioms

We tested three translation methods: (1) SIA, (2) LIA, and (3) Direct Translation. For the EN -> ZH, ZH -> EN, and HI -> EN we evaluated a random subset of 500 sentences and for the EN -> UR we evaluated on 216 sentences. The Urdu idiom dataset was limited because we only translated the idiomatic sentences that had corresponding English and Urdu idioms. All methods were translated with GPT-3.5-Turbo and GPT-4o. For all translations, we set the temperature to 0.7. All examples in the table are for English -> Chinese translation.

**SIA Method** As shown in Figure 2 above, in the Semantic Idiom Alignment method, we extracted idioms from sentences and searched for their meanings in the data. Using SentenceTransformers paraphrase-MiniLM-L6-v2, we generated embeddings for English meanings which are vector representations. These non-zero vectors capture the semantic meaning of the phrases. We then compare them with target language idioms using cosine similarity with a threshold of 0.7 to find the best match. Cosine similarity works by calculating the cosine of the angle between two vectors. After each idiomatic meaning is converted to vectors in the previous step, cosine values are calculated between -1 and 1, based on how semantically similar the meanings are. We chose a threshold of 0.7 because, through repeated trials, we found it to be the best at providing the most matches with minimal inaccuracy. If no match was found, we used the English meaning for translation. For the idioms that did find a match, we prompted GPT4o to choose/confirm an idiom if the lookup method found corresponding idioms in the dataset. We then translate the sentence while providing the corresponding target language idiom, as shown in Table 2 below.

Table 2: SIA Method Prompts

| SIA method CoT Prompt 1 | *You are a linguistic researcher on idioms and are good at Chinese and English. Choose the best Chinese idiom that matches the following English idiom and its definition. English idiom: '[English idiom]' English definition: '[English definition]' Here are some options: '[Chinese idioms]'.* |
|---|---|
| SIA method CoT Prompt 2 | *'[Chinese idiom 1]' (0.78), '[Chinese idiom 2]' (0.72), '[Chinese idiom 3]' (0.70), '[Chinese idiom 4]' (0.72). Please select the most relevant Chinese idiom and provide a brief explanation.* |
| SIA method CoT Prompt 3 | *'[English idiom]' means '[Chinese idiom]'. Given the above knowledge, translate this sentence to Chinese: '[English sentence]'.* |

**LIA method** For the LLM-based Idiom Alignment method, we first use GPT 4o to generate corresponding idioms in the target language that have the same meaning as the idiom in the original language. We give an option for the model to find up to 3 matches, specifically clarifying that it is acceptable not to find any match at all to minimize hallucinations. Then we prompt the model again to choose the best match from the top 3. As shown in Table 3 below, we do this to stay consistent with the GPT confirmation step performed in the SIA method. Lastly, we prompt the model to use the top LLM-generated idiom when translating the sentence. The key difference between the LIA method and the SIA method is that the target language idioms in the LIA method are generated by an LLM rather than extracted from a knowledge base.

### 3.3 Evaluation method

To evaluate the translations, we compared the original sentence and the translated sentence. We used both GPT4 and GPT4o as well as human evaluations. As shown in Table 4, the focus of the evaluation depended on whether the model was instructed to use a specific idiom in the translation. If there was an idiom in the translated sentence we instructed the model to focus on the idiom counterpart, but if there was not an idiom in the translated sentence we instructed the model to focus on whether the idiom's figurative meaning was maintained. We did this to ensure that the evaluation prompt

Table 3: LIA Method Prompts

| LIA Method CoT Prompt 1 | *You are a linguistic researcher on idioms and good at Chinese and English. You'll be provided an English idiom and your task is to: 1. First provide the definition of the idiom: '[Placeholder for English idiom]'. 2. Then find the three most similar Chinese idioms to the English idiom: '[English idiom]', and make sure to maintain context and cultural nuances. Follow these instructions: 1. If you cannot find three similar Chinese idioms, return as many as you can find. 2. If no Chinese idiom has the same meaning, only define the English idiom. 3. For good matches, respond with the Chinese idiom without pinyin and ensure it is an actual idiom, not a literal translation.* |
|---|---|
| LIA Method CoT Prompt 2 | *You are a linguistic researcher on idioms and good at Chinese and English. Choose the best Chinese idiom matching the English idiom and its definition. English idiom: '[English idiom]' English definition: '[English definition]' Options: Chinese idiom 1: '[Chinese idiom 1]' Chinese idiom 2: '[Chinese idiom 2]' Chinese idiom 3: '[Chinese idiom 3]'. Select the most relevant Chinese idiom and provide a brief explanation.* |
| LIA Method CoT Prompt 3 | *You are a linguistic researcher on idioms and are good at Chinese and English. '[English idiom]' means '[Chinese idiom]'. Given the above knowledge, translate the following sentence to Chinese: '[English sentence]'* |

was fairly tailored for each translation. We also set the temperature to 0.1 for the evaluations so there is less randomness. For the human evaluations, we provided the evaluators with the original sentences, the meaning of the idiom, and the 2 translated sentences anonymously. (GPT 3.5 and GPT-4o). We then gave them the exact task prompt and evaluation criteria that we gave the evaluation models. Every translation received a score from 1-3. Human evaluators were volunteers who were fluent in the language they evaluated. Evaluators didn't receive compensation. The specific task prompts and evaluation criteria are outlined in the table below:

Table 4: Evaluation Prompts

| **Task Prompt (No idiom):** Evaluate the idiom translation in the given Chinese translation of an English sentence. Focus on the idiom's figurative meaning. |
|---|
| **Task Prompt (With idiom):** Evaluate the idiom translation in the given Chinese translation of an English sentence. Focus on the idiom's counterpart in the translated language. |
| **Evaluation Criteria:** 1 point: Ignores, mistranslates, or only translates the literal meaning of the idiom. 2 points: Conveys basic figurative meaning but may lack refinement or have minor imperfections. 3 points: Exceptional translation, accurately conveying figurative meaning, context, and cultural nuances. |
| **Test Data:** Evaluate the following translation: English sentence: <source> Idiom in the English sentence: <idiom> Chinese translation: <translation> Evaluation (score only): <score> |

## 4 Results

The evaluations from our testing presented below reveal the performance of different models for translating idiomatic expressions from English to Chinese, Chinese to English, English to Urdu, and Hindi to English. The GPT-4o translations, expectantly, outperformed the GPT3.5-Turbo translations. Regarding the translation model, the GPT-4o evaluations consistently score the translations lower than the GPT4 evaluations; the evaluation done by GPT-4o matched more closely with the human evaluations. Using a binary correlation we found that the GPT4o score matched the human evalua-tion score 65% of the time while the GPT4 score only matched 53% of the time. The superior GPT4o model was more critical of the idiom translations than GPT4, making it a more human-like evaluation. As shown in Table 11, although the LLM evaluations typically did not score the SIA method the highest, the GPT-4o SIA method scored the highest on the human evaluations( which were evaluated using the same criteria as the LLM), making it a promising and viable method.

### 4.1 English -> Chinese and Chinese -> English

For the SIA EN->ZH translation, 238 idioms did not find a match, and 262 did, with results shown in Table 5. For SIA ZH->EN, 386 idioms did not find a match and 114 did, with results shown in Table 6. Despite the dataset not being designed for idiom-to-idiom correlation, the method still found success in translation. The translations that did not find an idiom not only scored better than the translations that did find an idiom in the LLM evaluations of SIA, but also the LLM evaluations of the LIA, as shown in Table 7 and Table 8. However, the human evaluations show that the translations that did find an idiom were mostly better translations. This suggests that the LLM is not adequately equipped to assess the accuracy of translations that contain idioms as it prefers the usage of the figurative meaning in the translation over a corresponding idiom. This is likely why the LLM evaluations also favored direct translation as it was better able to assess the accuracy of an idiom -> meaning translation rather than an idiom -> idiom translation, which can be better seen in Figure 2 below. Occasionally the SIA method fell short when the meanings were semantically similar but not the same. For example, "having extremely poor or no vision" ("blind as a bat") was paired with "having small and narrow vision; lacking in foresight ("目光如豆"). These two idioms being considered semantically similar is reasonable but the differences in the meaning account for the poor idiomatic translation. The majority of SIA method usages are successful such as pairing "to remain silent or keep a secret" ("zip one's lips") with "keep one's lips sealed, remain silent" ("缄口不言"). The LLM-Generated Idiom method scored lower likely due to the model not producing good idiom translations in the first place compared to the SIA method. The outputted idioms were very sensitive to the prompt as slight variations in the prompt led to varying idioms which could be a reason for the method's worse performance. The direct translation performed surprisingly well because for simple idioms such as "quality time" it was able to successfully translate it without additional information, as shown in Table 9 and Table 10.

### 4.2 English -> Urdu

For the EN -> UR sentences, 48 sentences were found in the English sentences dataset while 168 were generated by GPT4o. As shown by Table 12, the low resource language results showed the SIA underperforming. We attribute this to the LLM evaluations previously favoring the usage of the figurative meaning in the translation rather than a corresponding idiom, which is especially true here because, for the Urdu idioms dataset, we had a 1:1 correspondence for idioms. This means that all 216 English idioms had exactly one matching Urdu idiom. This was the case because the Urdu idiom dataset only had 216 idioms that matched an idiom in the English idiom dataset. Following the trend of the previous translations we hypothesize that human evaluations would show even more positive results for the SIA method.

### 4.3 Hindi -> English

Similarly, for the HI -> EN translation, the LIA method and direct translation were favored by the LLM evaluations, as shown in Table 13. As shown in Table 14, the human evaluations for the HI -> EN translations show the LIA method performing the best for the GPT3.5-turbo translations and the direct translation performing the best for GPT-4o translations, with the SIA method only scoring slightly worse. Our SIA method and LIA idiom method prove to be viable, promising methods by being on par and even at times exceeding the direct translation. GPT-4o's direct translations were successful because they provided simple translations that captured the meaning of the original sentence, even though they lost the idiomatic essence, whereas our methods preserved that idiomatic essence. Overall, both the SIA method and LIA method had the most complete translations when the corresponding idiom that was chosen was high quality, but direct translation still proved to be adequate at times.

Table 5: SIA method evaluations (Zh→En)

| Translation Model | Evaluation Model | Cosine Evaluations | Non-Cosine Evaluations |
|---|---|---|---|
| GPT 3.5 | GPT 4.0 | 2.4561 | 2.7798 |
| GPT 3.5 | GPT-4o | 1.7719 | 1.8964 |
| GPT-4o | GPT 4.0 | 2.5439 | 2.8938 |
| GPT-4o | GPT-4o | 2.0526 | 2.2668 |

Table 6: LIA method evaluations (En→Zh)

| Translation Model | Evaluation Model | Idiom:No Idiom Ratio | No Idiom Eval. | Idiom Eval. | Total Avg Score |
|---|---|---|---|---|---|
| GPT 3.5 | GPT 4.0 | 486:14 | 2.8571 | 2.7840 | 2.786 |
| GPT 3.5 | GPT-4o | 486:14 | 2.4286 | 2.3786 | 2.380 |
| GPT-4o | GPT 4.0 | 486:14 | 2.8571 | 2.7901 | 2.792 |
| GPT-4o | GPT-4o | 486:14 | 2.6429 | 2.4403 | 2.446 |

Table 7: LIA method evaluations (Zh→En)

| Translation Model | Evaluation Model | Idiom:No Idiom Ratio | No Idiom Eval. | Idiom Eval. | Total Avg Score |
|---|---|---|---|---|---|
| GPT 3.5 | GPT 4.0 | 494:6 | 2.8333 | 2.6356 | 2.638 |
| GPT 3.5 | GPT-4o | 494:6 | 2.0000 | 1.9291 | 1.930 |
| GPT-4o | GPT 4.0 | 494:6 | 2.8333 | 2.8036 | 2.804 |
| GPT-4o | GPT-4o | 494:6 | 2.3333 | 2.3016 | 2.302 |

Table 8: Direct translation evaluations (En→Zh)

| Translation Model | Evaluation Model | Average Score |
|---|---|---|
| GPT 3.5 | GPT 4.0 | 2.776 |
| GPT 3.5 | GPT-4o | 2.322 |
| GPT-4o | GPT 4.0 | 2.898 |
| GPT-4o | GPT-4o | 2.638 |

Table 9: Direct translation evaluations (En→Zh)

| Translation Model | Evaluation Model | Average Score |
|---|---|---|
| GPT 3.5 | GPT 4.0 | 2.776 |
| GPT 3.5 | GPT-4o | 2.322 |
| GPT-4o | GPT 4.0 | 2.898 |
| GPT-4o | GPT-4o | 2.638 |

Table 10: Direct translation evaluations (Zh→En)

| Translation Model | Evaluation Model | Average Score |
|---|---|---|
| GPT 3.5 | GPT 4.0 | 2.754 |
| GPT 3.5 | GPT-4o | 2.014 |
| GPT-4o | GPT 4.0 | 2.922 |
| GPT-4o | GPT-4o | 2.452 |

Table 11: Human evaluations

| Translation Direction and Model | Method Used | Average Score |
|---|---|---|
| EN → ZH GPT3.5 | SIA | 2.147 |
| EN → ZH GPT3.5 | LIA | 2.180 |
| EN → ZH GPT3.5 | Direct Translation | 2.245 |
| ZH → EN GPT3.5 | SIA | 2.428 |
| ZH → EN GPT3.5 | LIA | 2.142 |
| ZH → EN GPT3.5 | Direct Translation | 2.523 |
| EN → ZH GPT4o | SIA | 2.409 |
| EN → ZH GPT4o | LIA | 2.180 |
| EN → ZH GPT4o | Direct Translation | 2.360 |
| ZH → EN GPT4o | SIA | 2.761 |
| ZH → EN GPT4o | LIA | 2.333 |
| ZH → EN GPT4o | Direct Translation | 2.619 |

Table 12: Low resource language evaluations (En→Ur)

| Translation Model | Evaluation Model | Average Score |
|---|---|---|
| **SIA** | | |
| GPT 3.5 | GPT 4.0 | 2.425 |
| GPT 3.5 | GPT-4o | 2.000 |
| GPT-4o | GPT 4.0 | 2.430 |
| GPT-4o | GPT-4o | 2.203 |
| **Direct Translation** | | |
| GPT 3.5 | GPT 4.0 | 2.481 |
| GPT-4o | GPT 4.0 | 2.879 |
| GPT 3.5 | GPT-4o | 1.837 |
| GPT-4o | GPT-4o | 2.629 |

Table 13: Low resource language evaluations (Hi→En)

| Translation Model | Evaluation Model | Average Score |
|---|---|---|
| **SIA** | | |
| GPT 3.5 | GPT 4.0 | 2.522 |
| GPT 3.5 | GPT-4o | 1.968 |
| GPT-4o | GPT 4.0 | 2.478 |
| GPT-4o | GPT-4o | 2.036 |
| **Direct Translation** | | |
| GPT 3.5 | GPT 4.0 | 2.568 |
| GPT 3.5 | GPT-4o | 1.888 |
| GPT-4o | GPT 4.0 | 2.710 |
| GPT-4o | GPT-4o | 2.232 |
| **LIA** | | |
| GPT 3.5 | GPT 4.0 | 2.518 |
| GPT 3.5 | GPT-4o | 2.180 |
| GPT-4o | GPT 4.0 | 2.484 |
| GPT-4o | GPT-4o | 2.234 |

## 5 Limitations

Although the results of the SIA method have been promising thus far, there have been limitations in our work that prevented the method from being an even bigger success.

**Finite amount of idioms** As stated earlier in the LLM-generated idioms method, we could generate a corresponding idiom in the target language for nearly every original idiom. This yielded a much higher percentage of idioms that found a match, even if they were not all perfect matches. However the IdiomKB datasets, which were used in the SIA method, were composed of English and Chinese idioms without a 1:1 correspondence. There were 8,643 Chinese idioms and 3,990 English idioms. As a result, only about 1/2 of the idioms had a match in the SIA method. Had there been a comprehensive dataset that had both the English idiom and its corresponding Chinese idiom, the method would have been much more effective, which we leave to future work. Further, we leave the expansion of the knowledge base to more low-resource languages as well as exploration of more sophisticated ways to measure semantic similarity that cosine similarity for future work.

**Inferior GPT evaluation** GPT evaluation does not always strongly mimic human evaluation, especially for Urdu translation, where we lacked access to an Urdu human evaluator.

## 6 Potential Risks

Although relatively risk-free, some risks associated with translation can come to fruition if left overlooked. Data bias and representation issues within the knowledge base could lead to culturally insensitive or offensive translations. Along the same line of reasoning, language is always evolving, which is why it is important that the knowledge base remains up-to-date, and as comprehensive as possible. If it fails to fit such criteria, misunderstandings could arise, which in important contexts, such as legal, medical, or diplomatic communications could create dire situations.

## 7 Conclusion

In this paper, we presented advancements in translating idiomatic expressions using LLMs. We evaluated two methods, Semantic Idiom Alignment, and LLM-based Idiom Alignment, using Direct Translation as a baseline. Our findings indicate that the SIA method is particularly effective in preserving idiomatic integrity and achieving higher translation fidelity. Despite sometimes yielding worse results than other methods, the SIA method proved to be an effective and viable option. LIA performed well but fell short compared to the SIA, while Direct Translation often missed idiomatic nuances. Human evaluations confirmed the effectiveness of the Cosine Similarity Look-up method, emphasizing the need for context-aware translations. We believe our methods to be very generalizable to other languages if there are adequate datasets. Our approach is robust as is compatible and remains effective across languages. The impact of this technology can be proven significant when used to enhance communication through more accurate and culturally resonant translations of literary and educational materials. By making literary works more accessible, this research can help bridge cultural gaps and promote cross-cultural literacy and education globally. It profoundly impacts literary and educational communities by preserving the original tone and style of literary works, allowing readers worldwide to experience texts as intended. By enhancing LLMs to maintain the style and tone of messages across languages, we acknowledge the crucial role idioms play in communication and how they can express authors' intent in their work, something that is often lost with direct translation from two languages.